\documentclass[letterpaper, 10 pt, conference]{ieeeconf}  

\IEEEoverridecommandlockouts                              
\title{\LARGE \bf
Multi-Momentum Observer Contact Estimation for Bipedal Robots}

\author{J. Joe Payne, Daniel A. Hagen, Denis Garagić, and Aaron M. Johnson
    \thanks{This material is based upon work supported by Palladyne AI}%
    \thanks{J.J. Payne and A.M. Johnson are with the Department of Mechanical Engineering, Carnegie Mellon University, Pittsburgh, PA, USA. D. Garagić and D.A. Hagen are with Palladyne AI Corporation, Salt Lake City UT, USA. \texttt{jjpayne@andrew.cmu.edu}}%
}
\usepackage{amsmath}
\usepackage{mathtools}
\usepackage{amsthm}
\usepackage{amssymb}
\usepackage{accents}
\usepackage{cite}

\usepackage{algorithm}
\usepackage{algpseudocode}
\usepackage[bookmarks=true]{hyperref}
\hypersetup{%
  breaklinks,
  bookmarksnumbered=true,
  bookmarksopen=true,
  colorlinks=true,
  linkcolor=black,
  urlcolor=black,
  citecolor=black
}

\usepackage{pifont}
\newcommand{\xmark}{\ding{55}}
\newcommand{\cmark}{\ding{51}}

\newcommand{\revise}[1]{{\color{black}#1}}

\begin{document}
\maketitle
\thispagestyle{empty}
\pagestyle{empty}
\begin{abstract}
As bipedal robots become more and more popular in commercial and industrial settings, the ability to control them with a high degree of reliability is critical.
To that end, this paper considers how to accurately estimate which feet are currently in contact with the ground so as to avoid improper control actions that could jeopardize the stability of the robot. 
Additionally, modern algorithms for estimating the position and orientation of a robot's base frame rely heavily on such contact mode estimates. 
Dedicated contact sensors on the feet can be used to estimate this contact mode, but these sensors are prone to noise, time delays, damage/yielding from repeated impacts with the ground, and are not available on every robot.
To overcome these limitations, we propose a momentum observer based method for contact mode estimation that does not rely on such contact sensors.
Often, momentum observers assume that the robot's base frame can be treated as an inertial frame.
However, since many humanoids' legs represent a significant portion of the overall mass, the proposed method instead utilizes multiple simultaneous dynamic models.
Each of these models assumes a different contact condition.
A given contact assumption is then used to constrain the full dynamics in order to avoid assuming that either the body is an inertial frame or that a fully accurate estimate of body velocity is known.
The (dis)agreement between each model's estimates and measurements is used to determine which contact mode is most likely using a Markov-style fusion method. 
The proposed method produces contact detection accuracy of up to 98.44\% with a low noise simulation and 77.12\% when utilizing data collect on a hybrid humanoid/exoskeleton.
\end{abstract}
\section{Introduction}
Without explicit knowledge of which feet are in contact with the ground, a legged robot may take unreasonable or unsafe control actions. 
As an example, if a robot expects to step on the ground while its foot is still mid-air or if it expects to swing a leg while it is still in contact with the ground under some load, this mismatch between expectation and reality can result in the robot tripping and potentially falling.
Therefore, it is imperative that the current contact mode of a robot is estimated so that appropriate control actions can be taken.

\begin{figure}[tb]
    \centering
    \includegraphics[width = \linewidth]{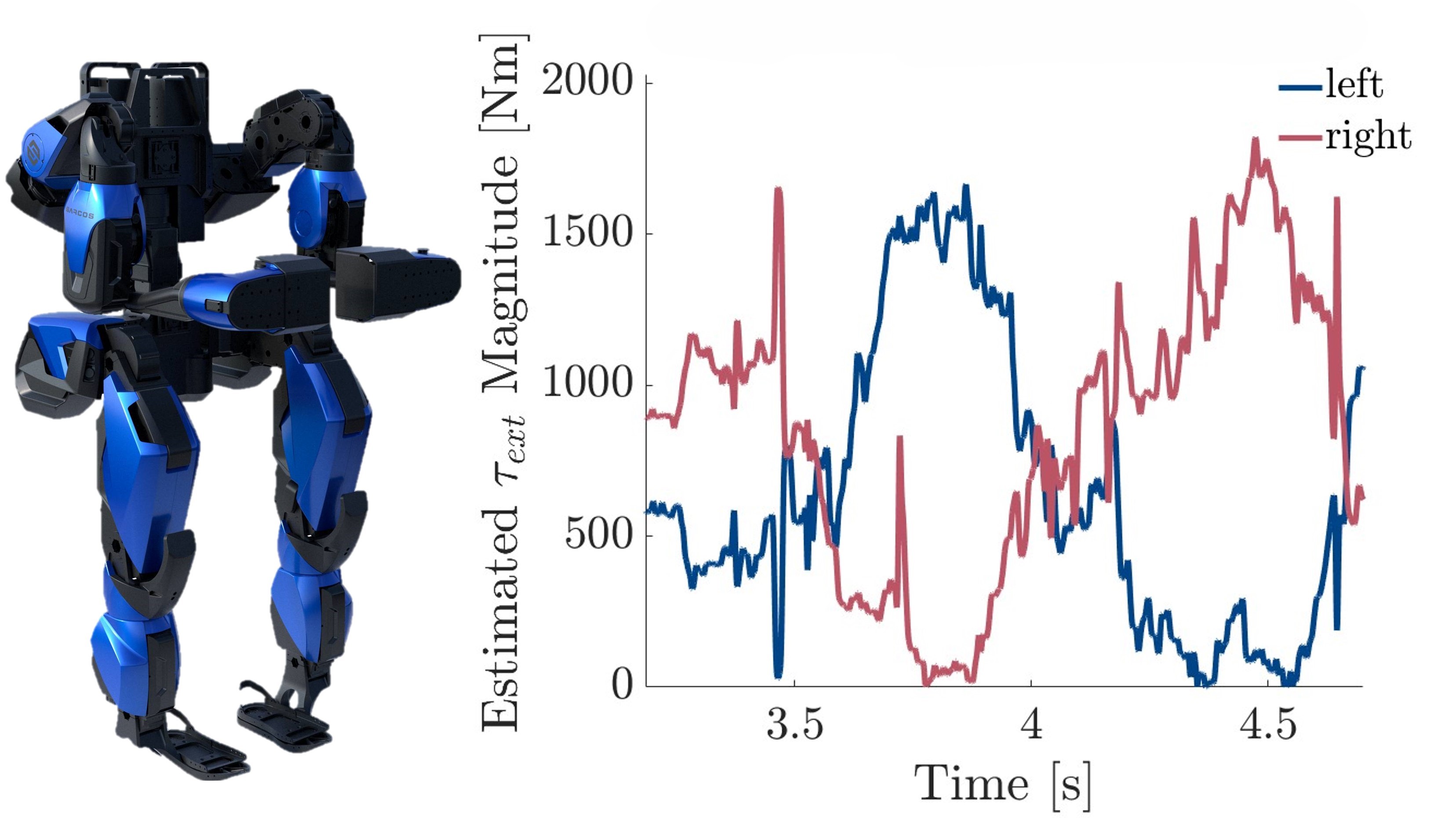}
    \caption[The bipedal system and its estimated external torques]{Left: The hybrid humanoid robot/exoskeleton on which we demonstrate these algorithms. Right: The estimated external torques on the system during a gait cycle using each separate momentum observer.}
    \label{fig:xo_system}
\end{figure}

Traditional methods for detecting the contact mode on legged robots make certain assumptions that often do not hold for bipedal (humanoid) robots.
These methods typically take one of two approaches.
First, a contact sensor, such as a switch or a force-torque sensor, is installed at the robot's foot so that contact detection becomes a problem of filtering or thresholding a single sensor's data \cite{bloesch2013state,rotella2014humanoid,hartley2018contactfactors,hartley2018contactstate,qin2020contactprobability}.
This method requires hardware changes and is not ideal for robots requiring robust sensing over long lifespans, as contact sensors are easily damaged due to the impulsive nature of repeated impact with the ground during locomotion.
The other method that is often used for contact detection is a thresholding method on estimated contact forces from joint torques \cite{camurri2017contact,yang2023cerberus}.
The issue with this method is that it assumes that the robot's body frame can be treated as an inertial frame and that the legs have significantly lower mass than the body.
This assumption does not hold in the case of many bipedal robots as the legs often represent a substantial portion of the robot's mass, such as the hybrid humanoid/exoskeleton shown in Fig.~\ref{fig:xo_system} which has close to 50\% of total body mass in \revise{the legs}, and therefore movement of the legs has an appreciable effect on the state of the body.

This work proposes a new contact estimation system that addresses these challenges through the use of multiple momentum observers that adopt different contact assumptions.
We avoid the assumption that the robot's body frame can be taken as an inertial frame by instead using the world frame as the inertial frame with an assumed stationary foot contact for the stance foot to reduce the order of the dynamics.
\revise{The proposed method focuses on bipedal robots as robots with more legs often have legs representing lower fractions of the total mass, enabling the use of the body as an inertial frame, or have more legs in contact at a given time, allowing quasi-static analysis.}
This assumption allows the 6 constraints at a given foot (in 3D) to remove the need to estimate the base link position, orientation, and velocities.
Then, a momentum observer \cite{deluca2005sensorless,deluca2006collision,haddadin2017collisions,bledt2018contact,yim2023proprioception} is applied to the constrained dynamics of the system in each contact mode.
By comparing the results of momentum observers on each of these assumptions, as well as monitoring the relative velocity between the feet, we obtain a probability of each contact condition (including dual support).

The proposed approach is thus a novel combination of reduced order dynamics, momentum observers, and Markov models which enables contact detection in large scale bipedal robots without (\emph{i}) direct contact sensors \revise{as in \cite{bloesch2013state,rotella2014humanoid,hartley2018contactfactors,hartley2018contactstate,qin2020contactprobability}}, (\emph{ii}) using the second derivative of any measurement or\revise{ precise knowledge of the body velocity as enabled by \cite{deluca2005sensorless,deluca2006collision,haddadin2017collisions,bledt2018contact,yim2023proprioception}}, or (\emph{iii}) assuming that the legs are much lighter than the body or moving much faster than the body \revise{as in \cite{camurri2017contact,yang2023cerberus}}.
The knowledge of contact states provided by this algorithm will enable the use of a variety of full body state estimators that rely on contact knowledge.

\section{Related Work}
There has been a wide array of research on state estimation for legged robotic systems.
In this section, we will primarily focus on floating base and contact estimation methods. 
Additionally, we will discuss the use of momentum observers for estimation of external contact forces.

\subsection{Floating Base Estimation}
Often, floating base estimators will assume that contact information is already provided.
To enable these floating base estimators, we will need to provide this information, which is not directly available in most bipedal robots.

In \cite{bloesch2013state}, an inertial measurement unit (IMU) is used as a process model in an extended Kalman filter (EKF) and the legged kinematics of a quadrupedal robot are used in the measurement update.
This work was then extended in \cite{rotella2014humanoid} to apply to humanoid robots.
In these algorithms, the contact information is used to update the process noise in the prediction step of the EKF.
In the case where a foot is not believed to be on the ground, its process noise is set to infinity (or a very large number) so that it is free to reposition.

In \cite{bloesch2013slippery}, IMU and kinematic data are fused through an unscented Kalman filter (UKF).
This method similarly rejects information from legs that are not detected to be in contact with the ground.
However, it additionally uses outlier rejection to ignore data from feet that may be slipping.
This outlier rejection could potentially be used to differentiate feet that are not in contact with the ground, however we propose a different method for contact detection.

Other works have used factor graph based methods for state estimation to fuse together different sensor modalities.
In \cite{hartley2018contactfactors,hartley2018contactstate}, forward kinematic constraints are added into a factor graph formulation alongside visual and IMU factors to estimate trajectories.
This work also requires a binary contact sensor or other means of determining active contact modes to determine which kinematic constraints are active between nodes on the factor graph.
Similar work was presented in \cite{wisth2019factorgraph}, which seeks to rely on contact classification less, but still uses estimates of external contact forces.
This was also extended in \cite{kim2022STEP} to include an additional foot velocity factor, but still utilized the kinematic constraint factors from the other works.

In \cite{yang2023cerberus}, the factor graph formulation is extended to simultaneously estimate parameters for the system that may be inaccurate such as link lengths.
While this improves over the other methods in many ways, it still requires a sensor to provide the current contact state to construct the kinematic constraints for the optimization problem.

\subsection{Contact Estimation}
Contact estimation is critical to our ability to estimate whole body state in legged robots.
There have been a variety of approaches to estimating the current contact conditions of a system.

Many methods have placed sensors directly on the feet to determine which contact modes are active.
In \cite{bloesch2013slippery,hartley2018contactfactors,hartley2018contactstate}, binary contact sensors are employed to determine the active contact modes for the floating base estimators.
Rather than a binary sensor, \cite{qin2020contactprobability} uses a force-torque sensor mounted on the feet to determine the active contact mode.
Similarly, \cite{yang2023multiimu} places additional IMUs on each of the feet as an additional means of sensing the active contact mode.
With the relatively slow walking gait of the bipedal robot we consider in this paper, the differences in velocities are not as drastic between contact modes as in quadrupeds with faster leg trajectories during walking gaits. Furthermore, the standard walking gait has the feet slow down before impact to minimize the impulse on the heavy legs. Therefore, while the feet of our system do have an IMU already on them, during slow walking they do not provide much information on the active contact mode.

Alternatively, many algorithms have sought to estimate contact modes through encoder and torque measurements in robot's legs \cite{johnson2010contact,wang2020contactlocalization}.
These internal sensors have also been used in methods such as the one proposed in \cite{bledt2018contact}, which utilizes a momentum observer. Our approach also uses a set of momentum observers as a part of the overall system, and we further discuss these systems in the following section.
However, these prior methods assume that the robot's legs are significantly lighter than the body in order to ignore inertial effects on the body. 
This assumption does not hold in the case of large bipedal robots as the legs represent a significant portion of the robot's mass.
Additionally, the method in \cite{bledt2018contact} relies on knowledge of the ground plane height as an additional means of determining contact probabilities.
Our method avoids making this assumption to remove the method's dependence on knowledge of the environment.

A comparison between each of these methods is presented in Table \ref{tab:comparison}.
Our proposed method is able to perform contact estimation without utilizing assumptions that do not hold for our large-scale bipedal robot and without modifying the hardware on the robot.
\begin{table*}[ht]
    \centering
\begin{tabular}{p{0.25\linewidth} | p{0.1\linewidth} p{0.1\linewidth} p{0.09\linewidth} p{0.08\linewidth}}
     & Rough Ground & Non-inertial Base Frame & No Hardware Modification & Low Speed\\
     \hline
     Proposed Method & \cmark & \cmark & \cmark & \cmark \\
     Leg-Momentum Observer \cite{bledt2018contact} & \xmark & \xmark & \cmark & \xmark \\
     Contact Sensor \cite{bloesch2013slippery,hartley2018contactfactors,hartley2018contactstate} & \cmark & \cmark & \xmark & \cmark \\
     Foot IMU \cite{yang2023multiimu} & \cmark & \cmark & \cmark & \xmark
\end{tabular}
\caption{Comparison between the proposed method and a variety of existing contact sensing methods.}
\label{tab:comparison}
\end{table*}

\subsection{Momentum Observers}
\label{mo_background}
One method for determining the current contact condition is to use a momentum observer.
Momentum observers, as introduced in \cite{deluca2005sensorless,deluca2006collision} and summarized in \cite{haddadin2017collisions}, utilize the generalized momentum of a system to estimate external contact forces. 
By examining momenta rather than accelerations, these methods have the benefits of avoiding costly inversions of generalized inertia matrices and the use of noisy estimates of joint accelerations from encoder data.
The difference between the expected and actual momenta at the end of each timestep are then taken to be the result of some externally applied torques.
Further explanation of the details of momentum observers can be found in Sec. \ref{mo_math}.

Recently, momentum observers have begun to see use in contact state estimation for legged robots.
In \cite{bledt2018contact}, the estimates from a momentum observer are fused with gait cycle information and the estimated foot height to determine the current contact mode. 
This method assumes a reasonable and steady gait cycle so as to determine the likelihood of contact based on clock timing, which does not always hold for more variable/dynamic walking behaviors.
Additionally, this method uses an estimate of the height of each foot as well as a measure of ground roughness as additional factors in determining contact probability.
As we seek to decouple floating base estimation and the contact state estimation, we cannot rely on estimates of the current foot height for contact mode estimation as they are typically derived from estimates of the centroidal state of the robot.
Alternatively, in \cite{yim2023proprioception}, a momentum observer is used to detect entanglements of a robot's legs with vines or other environmental disturbances during planned swing phases of a quadruped's gait cycle.
However, both of these methods implicitly assume that the robot's body can be taken as an inertial frame due to the relative masses of the bodies and legs of the robots, which does not hold for larger bipedal robots.

\section{Methods}
\subsection{Momentum observers}
\label{mo_math}
Many force estimation methods rely on acceleration data which is often noisy as it requires position data to be differentiated twice.
Alternatively, momentum observers avoid the use of these noisy acceleration estimates by instead utilizing the difference between the evolution of the expected momentum and the observed momentum of a system to estimate the external forces and torques acting on it. 

Consider a floating base (humanoid) robot with the following dynamic equations:
\begin{align}
    M\ddot{q} + C\dot{q} + G + A^{\top}\lambda = B^{\top}\left(\tau_{\text{mot}} + \tau_{\text{ext}}\right) \label{eq:mddq}
\end{align}

\noindent where $\{q, \dot{q}, \ddot{q}\}$ represent the generalized coordinates of the system (i.e., joint angles + base position/orientation) and their derivatives,  $M$ is the \revise{mass} matrix, $C$ is the Coriolis matrix, and $G$ is the gravity vector.
Modeled constraint forces ($\lambda$) act on the system through the constraint matrix $A$, while applied motor torques ($\tau_{\text{mot}}$) and unmodeled/external torques ($\tau_{\text{ext}}$) act through the selector matrix ($B^T\tau \equiv [0_{6\times1}^T~\tau^T]^T$) that simply ensures the floating base degrees-of-freedom (DOFs) are not actuated.
Trivially, it is possible to calculate the external torques with measurements/estimates of $q$ (and its derivatives), $\lambda$, and $\tau_{\text{mot}}$ utilizing the inverse dynamics approach.
But rather than relying on noisy acceleration data, we instead consider the derivative of the system's generalized momentum $p=M\dot{q}$:
\begin{align}
\begin{split}
\dot{p} & = \dot{M}\dot{q} + M\ddot{q} \\
& = \dot{M}\dot{q} - (C\dot{q} + G + A^{T}\lambda) + B^T\left(\tau_{\text{mot}} + \tau_{\text{ext}}\right)
\end{split}
\end{align}

\noindent Taking advantage of the identity $\dot{M} = C + C^{T}$, the dependence on the time derivative of the mass matrix can be removed to recover an equation for the rate of momentum that does not depend explicitly on joint accelerations:
\begin{align}
    \dot{p} &= C^{T}\dot{q} - G - A^{T}\lambda + B^T\left(\tau_{\text{mot}} + \tau_{\text{ext}}\right)\\
    &= \beta + B^T\tau_{\text{ext}}
\end{align}

\noindent where $\beta = C^{T}\dot{q} - G - A^{T}\lambda + B^T\tau_{\text{mot}}$. 
Using this formulation, now we maintain an estimate of the system's momentum, $\hat{p}$, and compare it to the measured momentum at each timestep to obtain estimates of the external torques $\hat{\tau}$, as in \cite{yim2023proprioception}:
\begin{subequations}
\label{basic_mo}
\begin{align}
p & = M\dot{q}\\
\hat{\tau} & = B^{-T}K_O(p-\hat{p})\\
\dot{\hat{p}} & =\beta + 
B^T\hat{\tau} 
\end{align}
\end{subequations}
where $K_O$ is an observer gain for the estimation of the external torques. 
Practically, this is implemented by discretizing the system's dynamics and considering the evolution of momentum estimates when $\dot{\hat{p}}$ is assumed constant over the discretized timestep.

\subsection{Constrained Dynamics}\label{sec:constrained_dynamics}
When a system is subject to holonomic constraints, such as flat-footed contact with the ground, the free dimensions are reduced by each independent constraint.
In these cases, we can utilize the linearization of the constraints at a given operating point to reduce the degrees of freedom of the system.
Consider the generalized dynamics from \eqref{eq:mddq}.
We can simultaneously select a set of reduced coordinates $y=Yq$ and enforce the constraints $A\dot{q} = 0$:
\begin{subequations}
\begin{align}
    & \begin{bmatrix} A \\ Y \end{bmatrix} \dot{q}  = \begin{bmatrix} 0 \\ I\end{bmatrix} \dot{y} \\
    & \dot{q} = \begin{bmatrix}
        A \\ Y
    \end{bmatrix}^{-1} \begin{bmatrix}
        0 \\ I
    \end{bmatrix} \dot{y} \label{eq:Hdef} \\
    \dot{q} = & H \dot{y}, \qquad H :=\begin{bmatrix}
        A \\ Y
    \end{bmatrix}^{-1} \begin{bmatrix}
        0 \\ I
    \end{bmatrix} 
\end{align}
\end{subequations}
where the choice of coordinates $y$ must be independent of the constrained directions such that the matrix in \eqref{eq:Hdef} is invertible. 
By using $\dot{q} = H\dot{y}$ and $\ddot{q} = H\ddot{y} + \dot{H}\dot{y}$, and noting that $H^T A^T = 0$, we can now rewrite \eqref{eq:mddq} as a reduced order (constrained) system that no longer explicitly depends on the constraint forces $\lambda$:
\begin{subequations}
\begin{align}
    \begin{aligned}
        &H^T MH\ddot{y} + \left( H^T CH + H^T M\dot{H}\right)\dot{y} \\
    &\hspace{2em} + H^T G + H^T A^T \lambda = H^T B^T(\tau_{\text{mot}} + \tau_{\text{ext}})
    \end{aligned} \\[1ex]
    \tilde{M}\ddot{y} + \tilde{C}\dot{y} + \tilde{G} = \tilde{\tau}_{\text{mot}} + \tilde{\tau}_{\text{ext}}\hspace{3.5em}
\end{align}
\end{subequations}
\noindent where,
\begin{subequations}
\begin{align}
    \tilde{M} &= H^T MH \\
    \tilde{C} &= H^T C H + H^T M \dot{H}\\
    \tilde{G} &= H^T G \\
    \tilde{\tau}_{\text{mot}} &= H^{T}B^T\tau_{\text{mot}} \\
    \tilde{\tau}_{\text{ext}} &= H^{T}B^T\tau_{\text{ext}}
\end{align}
\end{subequations}

\subsection{Multiple Model Observers}
\label{sec:multiple_mo}
For quadruped robots (with proportionately lighter limbs), the momentum observers described in \eqref{basic_mo} typically assume that the base frame of the robot can be treated as an inertial frame and thus only consider the dynamics of an individual leg. 
For larger bipedal robots, this assumption does not hold, so instead we utilize the ground as an inertial frame.
In order to do so, we consider the various \emph{contact-constrained} dynamics that arise from assuming only one foot in (flat) contact with the ground\footnote{It is important to consider flat contacts in 3D to fully constrain the system and to allow the constraint forces, $\lambda$, to be fully determined by the configuration of the robot (and its derivatives).}:
\begin{subequations}
\begin{align}
    H_i &= \begin{bmatrix}
        A_i \\ S_{\text{DOFs}}
    \end{bmatrix}^{-1} \begin{bmatrix}
        0 \\
        I
    \end{bmatrix} \\
    \dot{q} &= H_{i}\dot{y}_{i} \\[1ex]
    \ddot{q} &= H_{i}\ddot{y}_{i} + \dot{H}_{i}\dot{y}_{i} 
\end{align}
\end{subequations}

\noindent where $i\in\{l,~r\}$ represents the single support contact modes.
For each contact assumption, the reduced order dynamics are obtained (as in Sec. \ref{sec:constrained_dynamics}) by assuming the plant foot position/orientation is constant (i.e., $A_i\dot{q} = 0$) and selecting the joint positions as our reduced coordinates $Yq=S_{\text{DOFs}}q = y$.
\begin{align}
\tilde{M}_{i}\ddot{y}_{i} + \tilde{C}_{i}\dot{y}_{i} + \tilde{G}_{i} = \tilde{\tau}_{i,\text{mot}} + \tilde{\tau}_{i,\text{ext}}
\end{align}
Using these constrained system dynamics, we can separately maintain a momentum observer like those in \eqref{basic_mo} for each of the contact modes:
\begin{subequations}
\begin{align}
    p_i & = \tilde{M_i}\dot{y}_i\\
    \beta_{i} & = \tilde{C}_i^{T}\dot{y}_i - \tilde{G}_i + \tilde{\tau}_{i,\text{mot}} \\
\hat{\tau}_{i,\text{ext}} & = K_O(p_i-\hat{p}_{i})\\
\dot{\hat{p}}_{i} & = \beta_{i} +
\hat{\tau}_{i,\text{ext}} 
\end{align}
\end{subequations}

Intuitively, in situations where a leg is in single support and the other leg is in a swing phase and not impeded by its environment, then the estimated external torques $\hat{\tau}_{[l,r]}$ corresponding with that stance phase should be near zero.
We can utilize this information to inform our belief in which contact mode is currently active.
Additionally, when neither observer is estimating near zero external torques, it is likely that the robot is in a dual support phase (i.e., both limbs are experiencing "external" torques produced by contact with the ground).
Lastly, spikes in the estimated external torque for the active mode also represent a low latency means of detecting touchdown events.

\subsection{Velocity Constraints}
While the external torque estimates can quickly enable detection of touchdown events, they are not as sensitive to feet lifting off of the ground.
To fill in this gap, we additionally use the violation of a \emph{relative} velocity constraint as a means of detecting liftoff.
Specifically, during a dual support phase, the relative velocity between the two feet (calculated explicitly from joint measurements) should be zero, so when nonzero velocities are estimated it is likely that a liftoff event has occurred.
We can therefore use the relative motion between the two feet as a metric for how likely it is the system is in dual support:
\begin{align}
\begin{split}
    v_{\text{rel}} &= A_{lr}\dot{q} = (A_l - A_r)\dot{q}
\end{split}
\end{align}
\subsection{Model Fusion}
To combine the estimates from each of the momentum observers as well as the velocity constraint violation, we maintain an estimate of the likelihood of each contact condition.
To update this estimate, we utilize a Markov model based on how likely it is that each transition event has occurred.

To determine how likely a touchdown event is to have occurred from a swing-leg phase, a threshold value $\tau_{t}$ and a band value $\tau_{b}$ are defined to apply to a sigmoid activation function:
\begin{align}
    \pi_{i,\text{Dual}} = \text{sgm}\left(\frac{||\hat{\tau}_{i,\text{ext}}||_2-\tau_t}{\tau_b}\right)
\end{align}
\noindent where $\pi_{i,Dual}$ is the probability of transition from the given single support mode into dual support.

Similarly, when calculating the transition probability from dual support to single support, we apply a sigmoid function centered about $v_t$ with width $v_b$ to the estimated velocity constraint violation. 
Then we determine which stance leg is more likely by comparing the relative magnitudes of external torques estimated by the two momentum observers -- whichever has a lower estimate is more likely to be the active mode.
\begin{align}
    \pi_{\text{Dual},i} = \text{sgm}\left( \frac{v_{\text{rel}}-v_t}{v_b}\right)\frac{||\hat{\tau}_{j,\text{ext}}||_2}{||\hat{\tau}_{i,\text{ext}}||_2+||\hat{\tau}_{j,\text{ext}}||_2}
\end{align}
\noindent where, for each $i \in \{l,r\}$, $j\in\{l,r\}\backslash\{i\}$ and $\pi_{Dual,i}$ is the transition probability from dual support to the given single support mode.

The means and widths for each of these transition probability sigmoids can be tuned for a given system to achieve some desired performance characteristics.
By combining these transition probabilities, we are able to calculate a Markov transition model for each timestep to update the likelihood of each given contact condition.
We maintain the likelihood of each mode in vector form as:
\begin{align}
    \mathrm{P} =
    \begin{bmatrix}
        \mathrm{p}_{l} & \mathrm{p}_{r} & \mathrm{p}_{\text{Dual}}
    \end{bmatrix}^T
\end{align}
To update these probabilities, we construct the transition matrix $\Pi$ at each timestep based on the individual transition likelihoods:
\begin{align}
    \Pi = \begin{bmatrix}
        1-\pi_{l,\text{Dual}} & 0 &  \pi_{\text{Dual},l}\\
        0 & 1-\pi_{r,\text{Dual}} & \pi_{\text{Dual},r}\\
        \pi_{l,\text{Dual}} & \pi_{r,\text{Dual}} & 1 - \pi_{\text{Dual},l} - \pi_{\text{Dual},r}
    \end{bmatrix}
\end{align}
\noindent such that we can update the probabilities at each timestep by:
\begin{align}
    \mathrm{P}_{k+1} = \Pi\ \mathrm{P}_k
\end{align}

To select the active contact mode estimate, we take the most likely mode at a given time.
This may sometimes result in selecting an active mode that is estimated to have less than a 50\% likelihood.
In practice, there are many ways to handle this information.
The likelihoods themselves could be fed into other estimation methods as a means of representing how much certainty should be associated with dynamic constraints from a contact mode.
Alternatively, an additional mode that represents an uncertain contact condition could be added in order to prevent switching when uncertainty is more likely.
For the work proposed here, the contact mode estimate is only allowed to transition to a new contact mode once a threshold of belief in a given mode is crossed (e.g., $>50$\% likelihood for a given mode is required to switch to that mode).

\section{Results}
\subsection{Simulated Walker}
\begin{figure}[tb]
    \centering
    \includegraphics[width = 0.5\linewidth]{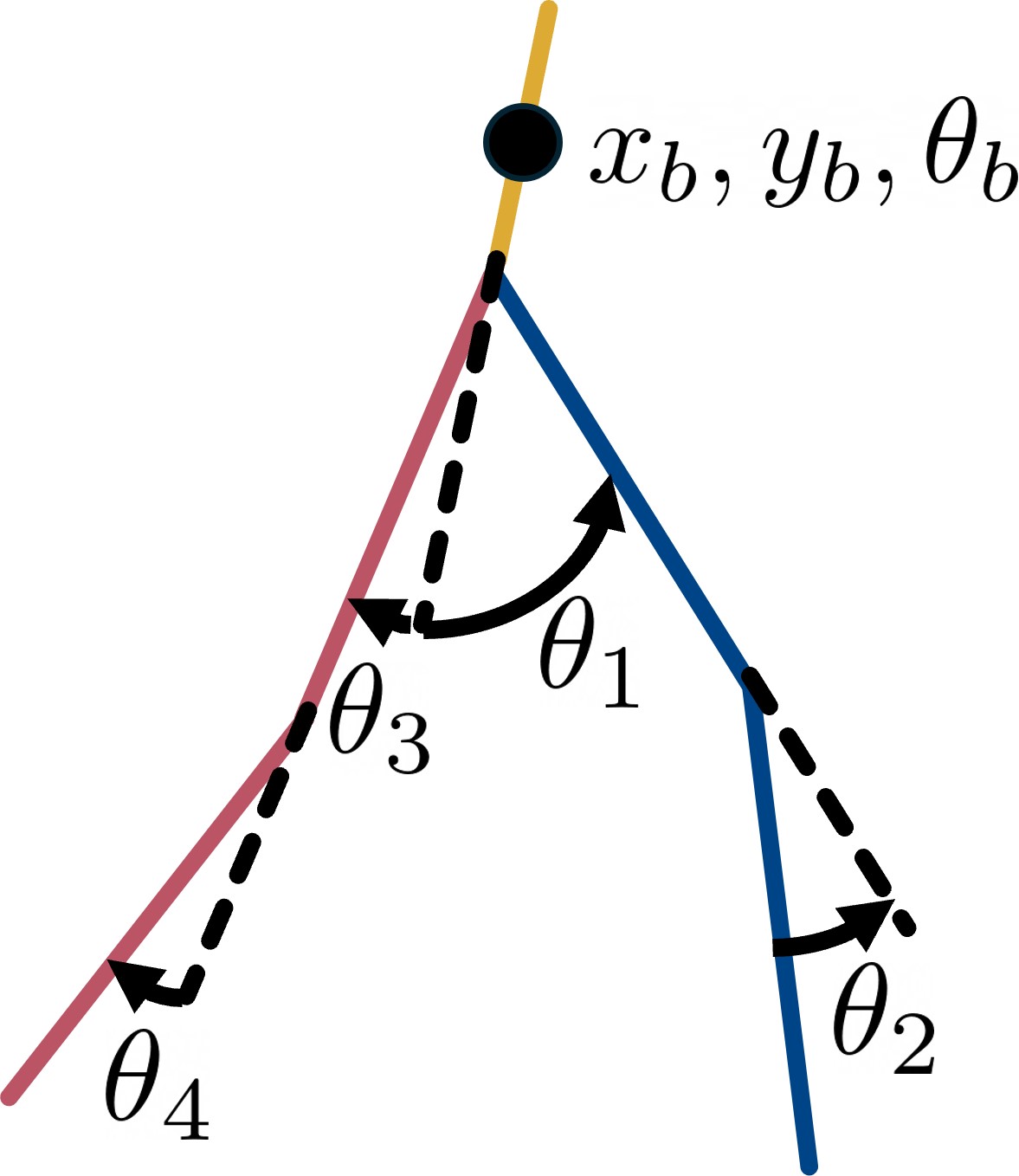}
    \caption{The five link walker model we use in simulation to evaluate the algorithm.
    The floating base location is defined as a position \revise{($x_b,y_b$)} and angle \revise{($\theta_b$)} relative to the vertical.
    The link angles are defined relative to straight hip \revise{($\theta_1,\theta_3$)} and straight knee \revise{($\theta_2,\theta_4$)} positions.}
    \label{fig:five_link_system}
\end{figure}

In this section we present the results of estimation performed on a planar five link walker simulated in Drake \cite{drake}.
This walker was modeled to have significant mass distribution in its legs -- each leg weighs twice as much as the base link -- to ensure that the assumption that the floating base is an inertial frame does not hold.
The model used in this system is shown in Fig.~\ref{fig:five_link_system}.
Each link has a mass of 1 kilogram with a uniform mass distribution.
The system has seven degrees of freedom, the position and orientation of the base link are three unactuated degrees of freedom, and the angle of each of the joints in the legs -- left and right hips and knees -- are four actuated degrees of freedom.
The state representation of the system is:
\begin{align}
    q = \begin{bmatrix}
        x_b & y_b & \theta_b & \theta_1 & \theta_2 & \theta_3 & \theta_4
    \end{bmatrix}^T
\end{align}
and the dynamics matrices $M$, $C$, and $G$ and the constraint matrix $A$ are calculated using Lagrangian dynamics.
 
The bipedal walker is run through a left-right gait for these trials. 
Each momentum observer for this simulated system assumes that a foot is in point contact with the ground, which provides two constraints to the system.
Using the constrained dynamics, as discussed in Sec.~\ref{sec:multiple_mo}, we are able to remove the dependence of the estimator on the two body position variables, so the measurements used are the orientation of the base and each of the joint angles.

\begin{figure}[tb]
    \centering
    \includegraphics[width = \linewidth]{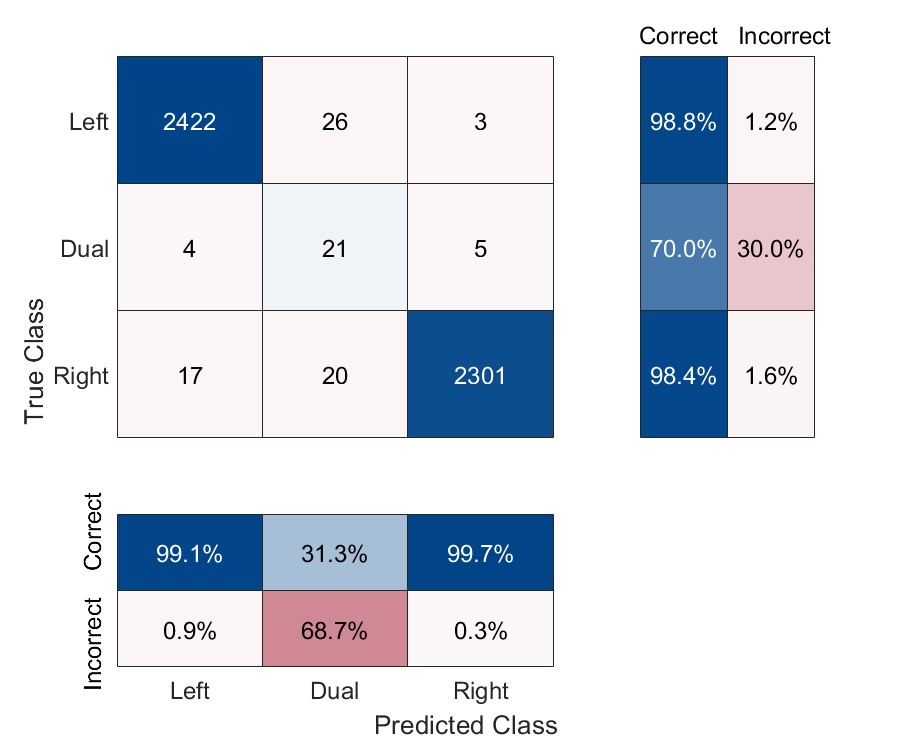}
    \caption{A confusion matrix for the mode estimation from the simulated five link walker over a five second trajectory with a signal to noise ratio of 70dB.
    The values on the diagonal in blue represent correct estimations and the values off the diagonal in red represent incorrect estimations. 
    The bottom and side contain the percentages of the time that a specific estimate is correct and that a specific mode is correctly estimated, respectively.}
    \label{fig:five_link_confusion}
\end{figure}

\begin{figure}[tb]
    \centering
    \includegraphics[width = \linewidth]{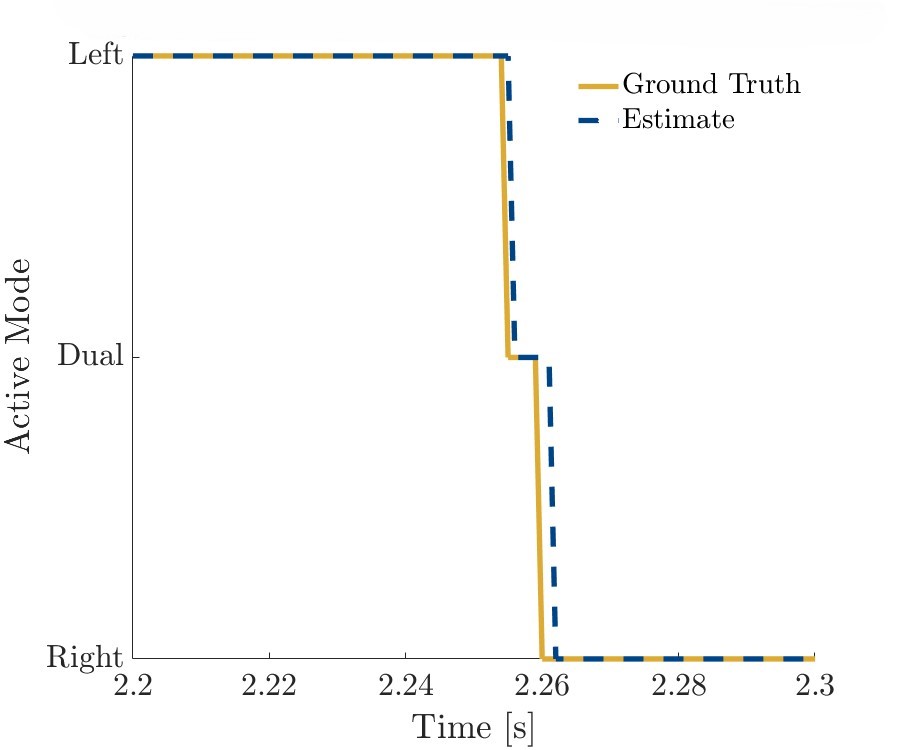}
    \caption{A comparison between the current contact mode estimate and the ground truth of the simulated five link biped system on a trial with a signal to noise ratio of 70dB. 
    The estimate is shown as a blue dashed line and the ground truth is shown as a solid yellow line.
    An inset highlights the short lag in detecting touchdown and liftoff at a stance transition event.}
    \label{fig:five_link_contact_est}
\end{figure}

For a single 5 second trial running at 1kHz using a signal to noise ratio of 70dB, the associated confusion matrix, which compares the estimated modes to the ground truth modes from the simulation, is shown in Fig.~\ref{fig:five_link_confusion}.
Similarly, a timeseries of the estimates for a single trial can be seen in Fig.~\ref{fig:five_link_contact_est}.
The associated mode accuracy for this system is 98.44\% with much of that error occurring in the initial timesteps while the momentum observers initialize.

\begin{figure}[tb]
    \centering
    \includegraphics[width = \linewidth]{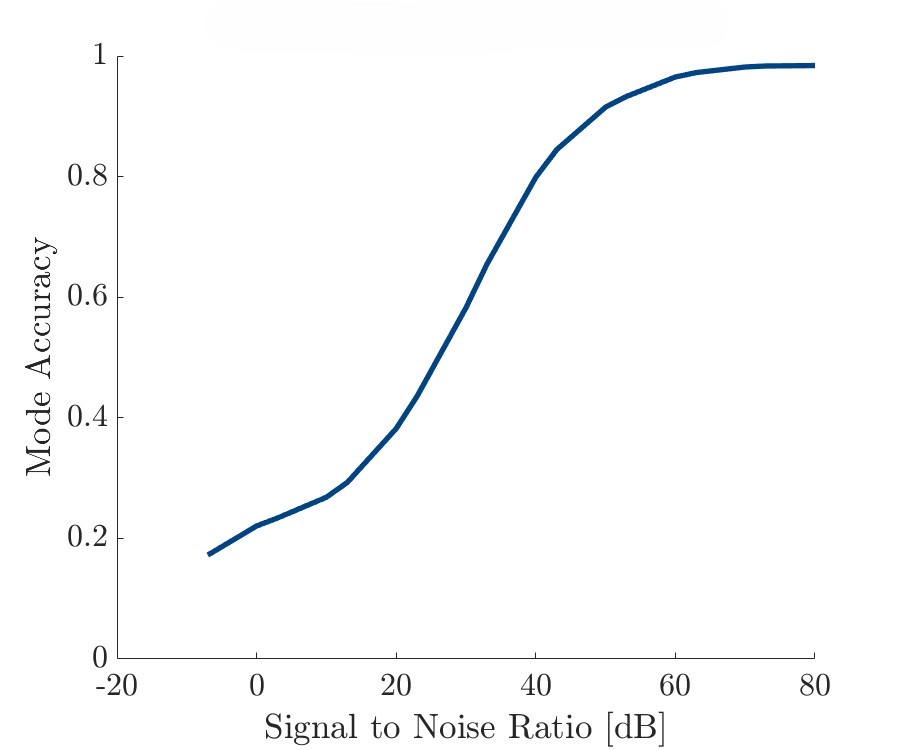}
    \caption{Estimation accuracy as compared to the signal to noise ratio of the system. With lower noise levels, we see very accurate mode estimation. However, as the momentum observer is sensitive to velocity noise, the performance degrades notably with higher noise levels.}
    \label{fig:error_noise_comparison}
\end{figure}

To better characterize the performance of the estimator across a variety of levels of noise in the system, we present a plot comparing the mode accuracy as a function of the signal to noise ratio (calculated as $10\log(\frac{\text{signal}}{\text{noise}})$) in Fig.~\ref{fig:error_noise_comparison} over 100 trials at each noise level with randomly sampled noise.
In this, we can see that when the noise in the system is minimal, a properly tuned estimator can reach near 100\% accuracy.
At higher noise levels, we see much lower accuracy.
As the momentum observer is highly dependent on accurate velocity estimation, the error increasing due to noise in the velocity makes sense.
This leads to neither observer estimating external torques near zero magnitude, so it is more likely to select dual support, leading to low accuracy.

\subsection{Exoskeleton Walker}
For this portion of the work, we utilize data collected from a hybrid humanoid/exoskeleton, which is shown in Fig.~\ref{fig:xo_system}.
This robot has a mass of 150kg and much of that mass is distributed in the limbs, so the assumption that the body link is an inertial frame cannot be adopted.
This data was collected during a trial in which the system walked as a humanoid on its own without a person inside controlling it.
The measurements used for this experiment are the measured joint torques (as opposed to the commanded), the orientation of the body via an IMU, and the joint positions via encoders.

\begin{figure}[tb]
    \centering
    \includegraphics[width = \linewidth]{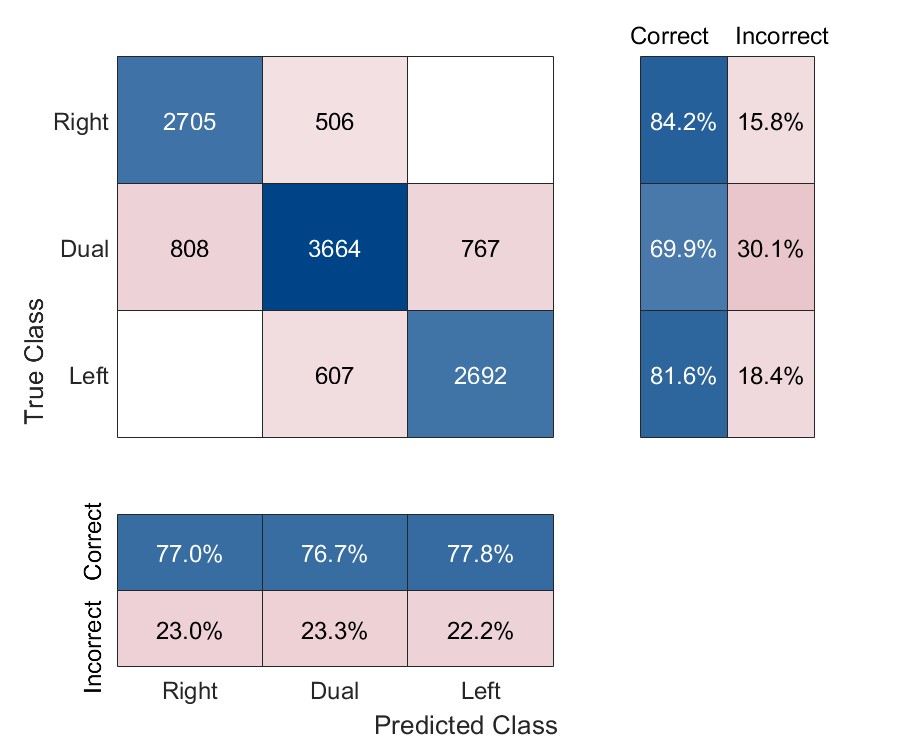}
    \caption{A confusion matrix for the mode estimation from the bipedal walker in hardware over an approximately twelve second trajectory.
    The values on the diagonal in blue represent times where the estimator and the video frame align and the values off the diagonal in red represent estimations where the two do not align. 
    The bottom and side contain the percentages of the time that a specific mode estimate aligns with the video and that a specific mode from the video is correctly estimated, respectively.}
    \label{fig:xo_confusion}
\end{figure}

\begin{figure}[tb]
    \centering
    \includegraphics[width = \linewidth]{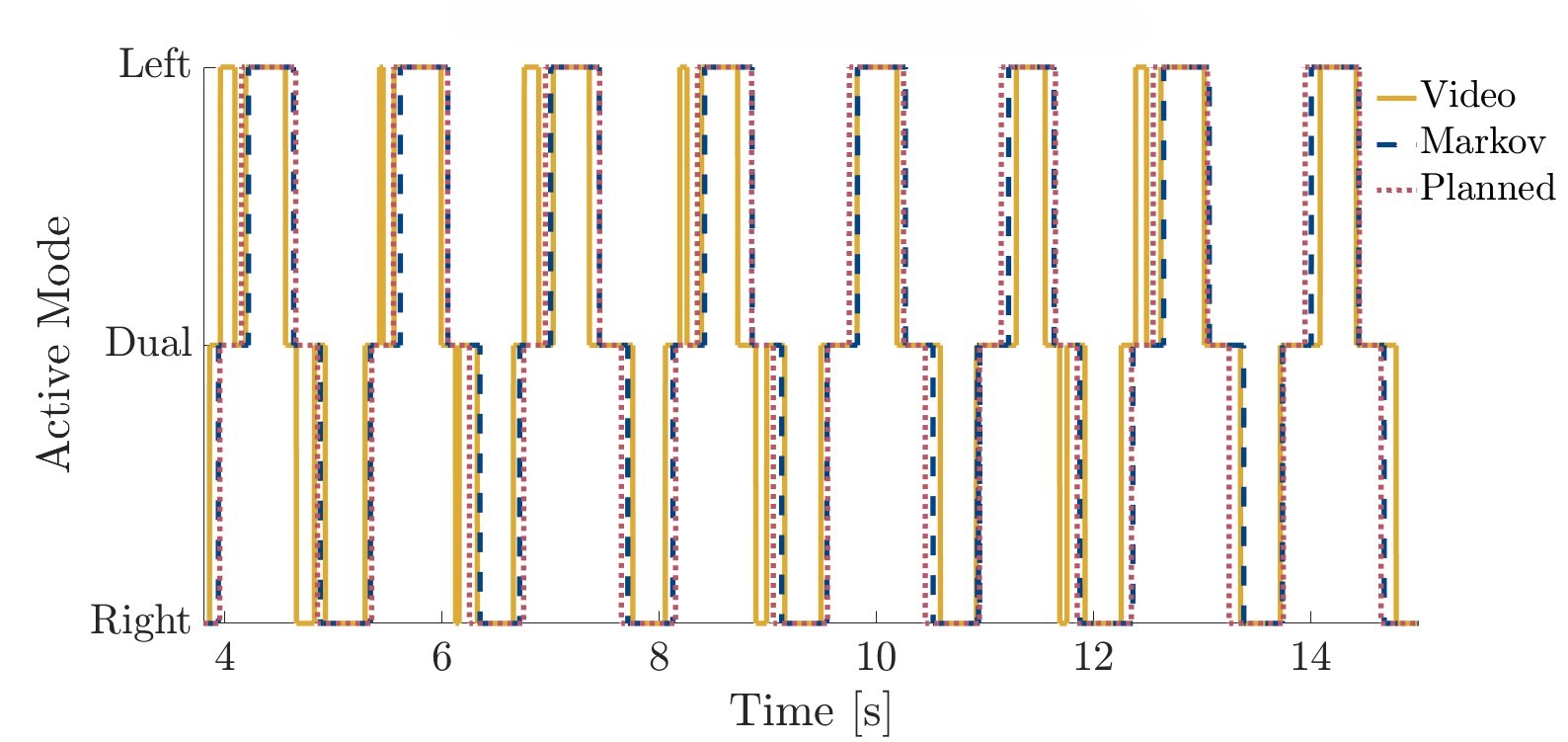}
    \caption{A comparison between the current contact mode estimate with the proposed method, the planned contact sequence, and the contact mode derived from the corresponding video. 
    The estimate is shown as a blue dashed line, the planned state is shown as a red dotted line, and the mode extracted from the video is shown as a solid yellow line.}
    \label{fig:xo_contact_est}
\end{figure}

All of the mode estimates on the system are compared against hand-labeled contact modes derived from a video of the corresponding trial.
A confusion matrix comparing the estimated mode to the mode extracted from the video is shown in Fig.~\ref{fig:xo_confusion}. 
The associated estimates from the trial are shown as a time series in Fig.~\ref{fig:xo_contact_est}.
The corresponding mode accuracy for this system is 77.12\%, compared to 70.15\% \revise{accuracy of the feedforward planned contact sequence from the motion planner}.
Directly utilizing the most likely contact mode for a given time without the probabilistic transition model results in a similar accuracy of 76.10\%, however it predicts twice as many transitions than truly occur compared to 27.5\% fewer than true transitions predicted by the transition model.
While this is not a perfect model of the transitions, it is preferable to have a stable estimate of the current contact state for control purposes.
Some of the misalignment can be explained by the difference in the estimation update rate and the camera frame rate.
Additionally, due to the hand-labeled nature of the ground truth, there are orientations in which it is difficult to discern the true contact conditions. However, the general timing and gait patterns align well between the estimate and the values derived from the associated video.
To improve the practical performance, the cutoff values for the sigmoids can be tuned to either over- or under-estimate dual support depending on the context in which the estimation is being used.
Additionally, \revise{to reach higher accuracy values for use in real time control}, this should be fused with more information, such as estimated ground heights from a visual or lidar sensor.
However, for this work we sought to make this estimation agnostic to the profile of the ground.

\section{Conclusion and Future Work}
In this work we show that through the use of multiple momentum observers we are able to obtain estimates of the active contact mode of a system through only proprioceptive means.
Using this method we are also able to avoid making the assumption that the base of the robot is an inertial frame.

We then demonstrate the effectiveness of this method by implementing it for both a simple simulated bipedal system and a large scale bipedal exoskeleton in hardware.
In simulation we show that the mode estimation accuracy is able to reach 98.44\% with low noise.
Additionally, in hardware we show that the estimator is able to track the active contact mode that we observe through our collected video during the trial more accurately than relying on the planned contact mode.

In combination with other sensor information, this method will allow robots without contact sensors in their feet to better estimate their current stance phases.
This will enable them to take more well informed control actions, leading to more robust gaits.

However, there is still room for extensions to this work.
So far, we have considered flat foot stance for each of the feet, limiting us to 3 contact modes.
This could be extended to include heel strike and toe-off modes to better understand the control authority of the system at a given time.

Additionally, this work assumes that there is no flight phase as the systems we considered are not capable of flight phases. 
This method could be modified to differentiate between situations where neither individual momentum observer agrees with the measurements when in dual stance or when in flight.

Similarly, we assume that any contact occurring corresponds to a foot touchdown event.
It is possible the robot will also be making contact or experiencing external forces and torques not due to foot touchdown.
An interesting extension would be to differentiate between ground contact and other external forces, such as tripping events as discussed in \cite{yim2023proprioception}.

While this method was designed specifically for bipedal systems, it could also be interesting to consider how these assumptions can be used to better understand the contact modes of quadrupedal and other systems.
While the mass distribution of those systems does make the assumption of taking the floating base as an inertial frame more reasonable, it still doesn't hold perfectly.
A difficulty with this extension would be considering the combinatorial nature of the contact modes.
\addtolength{\textheight}{-18.0cm}

\bibliographystyle{IEEEtran}
\bibliography{references}
\end{document}